# CASL-VAE: Learning Structured Latent Variables from Unpaired Data for Semi-supervised Clustering and Paired Sample Generation

Sai Spandana Chintapalli [1 2] Pratik Chaudhari [3] Christos Davatzikos [1 2 4]

## Abstract

Quantifying variability in a target population relative to a reference population is central to many scientific and clinical problems (e.g., diseased vs. healthy). Yet, without paired data and in the presence of heterogeneous target variation, existing methods struggle to separate multiple modes of target-specific variation. We propose *CASL-VAE*, a deep contrastive latent variable model that learns structured latent generative factors from unpaired data. CASL-VAE factorizes variation into continuous common latent factors shared across populations and hierarchical salient latent factors that model target-specific heterogeneity as discrete subtypes and continuous within-subtype variation. Using variational inference, we show how approximate joint likelihood optimization over reference and target domains can be performed using unpaired data, providing a principled basis for paired-sample generation and cross-domain analysis. We validate CASL-VAE on semi-synthetic neuroimaging data, demonstrating improved subtype recovery and paired-sample generation compared to baseline clustering and generative models. We also validate its ability to reveal biologically plausible heterogeneity in Alzheimer's disease.

## 1. Introduction

Many scientific and applied problems involve understanding how a target population varies with respect to a reference population. For example, in medicine, researchers analyze pathological patterns in patient data relative to healthy controls (Fan et al., 2008). Similar comparative analyses arise in genomics, drug discovery, and socio-economic studies, where contrasts between groups are often more informative than absolute measurements (Romero et al., 2012; Soneson & Robinson, 2018; Akhoon, 2021).

In practice, such comparisons are challenging for three reasons: (i) variation in the target domain is often entangled with variation present in the reference domain, obscuring subtle target-specific effects (for example, natural biological variation in healthy individuals could be mistaken for disease-related changes in patients); (ii) target-domain variation is frequently heterogeneous (Nunes et al., 2020), reflecting latent subtypes or context-dependent factors (for instance, patients with the same diagnosis may have diverse underlying pathologies, or cell populations may respond differently to the same treatment); and (iii) paired observations across domains—where each target sample has a corresponding reference sample—are rarely available, preventing direct supervision for learning cross-domain relationships. As a result, existing methods often struggle to disentangle heterogeneous, target-specific variation in unpaired data (Marquand et al., 2016; Feczko et al., 2019).

Generative representation learning can help address these challenges by learning latent representations that model the data generation process (Bercea et al., 2025). In principle, such representations can differentiate between domains, capture heterogeneous subpopulations, and support downstream generative analyses (Kopf & Claassen, 2021). However, standard generative models applied to multi-domain data typically entangle shared and domain-specific factors, leading to unstable or difficult-to-interpret representations (Higgins et al., 2017; Chen et al., 2016). Contrastive Analysis (CA) methods aim to explicitly separate common variation shared across reference and target domains from salient variation unique to the target domain (Abid et al., 2018; Severson et al., 2019). Yet existing formulations often lack sufficient latent structure to capture multi-modal target heterogeneity or to preserve sample-level semantics across domains, limiting their ability to support downstream generative tasks such as paired data generation.

To address these limitations, we propose ***CASL-VAE*** *(Contrastive Analysis with Structured Latents via Variational Autoencoder)*, a generative framework for learning structured and disentangled latent representations from unpaired reference–target data (Figure 1(a)). CASL-VAE factorizes the latent space into (i) a **continuous common latent** that captures variation shared across domains, and (ii) a **hierarchical salient latent** that models heterogeneous, target-specific variation via discrete latent subtypes and continuous within-subtype factors. The hierarchical design, regularized using domain-level labels (reference vs target labels)[1], enables **unsupervised target subtype discovery**. This structured latent decomposition acts as an inductive bias that, together with sufficient constraints, enables approximate joint likelihood learning across domains using only marginal observations. As a result, CASL-VAE supports **paired-sample generation** and **interpretable cross-domain comparisons** despite the absence of paired training data.

[1]Center for AI and Data Science for Integrated Diagnostics, University of Pennsylvania [2]Department of Bioengineering, University of Pennsylvania [3]Department of Electrical and Systems Engineering, University of Pennsylvania [4]Department of Radiology, University of Pennsylvania. Correspondence to: Christos Davatzikos <christos.davatzikos@pennmedicine.upenn.edu>.

We evaluate CASL-VAE on neuroimaging data, where brain measurements exhibit substantial inter-individual variability and heterogeneous pathological patterns are difficult to isolate. Paired measurements across conditions (e.g., observations of a participant in both healthy and diseased state) are typically unavailable, making unpaired contrastive modeling both practically relevant and technically challenging. While our experiments use neuroimaging-derived **regional brain volumes**, the proposed framework is broadly applicable to other settings that require unpaired reference–target comparisons and modeling of heterogeneous target variation. Using data from the iSTAGING consortium (Habes et al., 2021), we validate CASL-VAE on semi-synthetic data with known ground-truth disease patterns and severity, and compare it to representative clustering and generative modeling baselines. We further apply CASL-VAE to a real Alzheimer's disease cohort to demonstrate the clinical relevance of the learned structured latent variables. Our **contributions** are threefold:

- We introduce CASL-VAE, a contrastive generative framework that learns disentangled shared and target-specific representations from unpaired data using structured latent variables.
- We propose a hierarchical salient latent decomposition that enables unsupervised discovery of target subtypes while preserving continuous within-subtype variation.
- We demonstrate that this structure enables interpretable downstream capabilities, including paired-sample generation and cross-domain comparison, and validate these properties empirically.

[1]In this paper, the term *semi-supervised* refers to supervision available only at the domain level (reference vs. target), without any annotations for target subtypes or paired samples.

## 2. Related work

**Probabilistic latent variable models** such as variational autoencoders (VAEs) aim to learn compact representations of complex data (Diederik et al., 2014), while conditional variants model data distributions conditioned on domain labels (Sohn et al., 2015). From a representation learning perspective, these models are primarily optimized for data generation. Without additional modeling assumptions or regularization (Higgins et al., 2017), their latent representations may entangle shared and domain-specific sources of variation, limiting interpretability and downstream utility in comparative analyses.

**Contrastive analysis** (CA) methods address this by explicitly aiming to separate variation that is shared across two datasets from variation that is enriched in a target dataset. Early CA approaches relied on linear factor models (Ge & Zou, 2016; Abid et al., 2018). More recent work extends CA to nonlinear settings using probabilistic latent-variable models (Abid & Zou, 2019; Severson et al., 2019), often with additional regularization to prevent information leakage and preserve the semantic meaning of the learned factors (Weinberger et al., 2022; Louiset et al., 2024). While this enables separation of shared and target-specific variation, it does not explicitly model heterogeneous structure within the target population.

In many real-world applications, however, separating two populations is not sufficient: the target population itself often exhibits meaningful heterogeneity that is critical to understand. For example, CA has been applied to single-cell RNA-seq (Weinberger et al., 2023; Jones et al., 2022) and neuroimaging (Aglinskas et al., 2022; Zheng et al., 2024) datasets, where target-specific variation is frequently multi-modal and reflects distinct biological subtypes or phenotypes that need to be identified. In these settings, heterogeneity is typically addressed post hoc, for example, by clustering the learned salient representations. However, when target effects are subtle or subtypes are partially entangled, such representations may not form well-separated clusters. This issue is further exacerbated in VAE-based CA models, which can be prone to posterior collapse, potentially conflating or collapsing multi-modal target variation unless additional structure is imposed.

**Clustering-based approaches** have also been proposed for capturing target heterogeneity (Gao et al., 2023). Standard clustering methods, however, do not leverage contrastive structure and therefore may produce clusters that are driven by shared variation rather than target-specific effects. Some recent methods cluster based on differences between target and reference populations (Chand et al., 2020; Yang et al., 2021). While effective for unsupervised subtype discovery, these approaches typically optimize discrete clustering objectives rather than learning reusable generative latent representations and therefore have limited interpretability.

**CASL-VAE** addresses these limitations by integrating contrastive analysis with structured hierarchical latent variables within a single generative framework. Designed for unpaired reference–target data, it enables the modeling of target subtypes and within-subtype variability. The framework supports data-driven, unsupervised discovery of target-specific heterogeneity while remaining interpretable, distinguishing it from many prior contrastive and clustering-based approaches that rely on post-hoc analyses or subtype labels.

## 3. CASL-VAE

### 3.1. Problem Setup

Given unpaired observations from two related domains, a target domain $\mathcal{X} = \{x_i\}_{i=1}^{N_x}$, $x_i \sim p(x)$ and a reference domain $\mathcal{Y} = \{y_j\}_{j=1}^{N_y}$, $y_j \sim p(y)$ where no correspondence is assumed between sample $i$ and $j$. Our goal is to learn latent representation that separates variation shared across these domains from heterogeneous, multi-modal variation specific to the target domain, enabling target-subtype discovery and cross-domain comparisons.

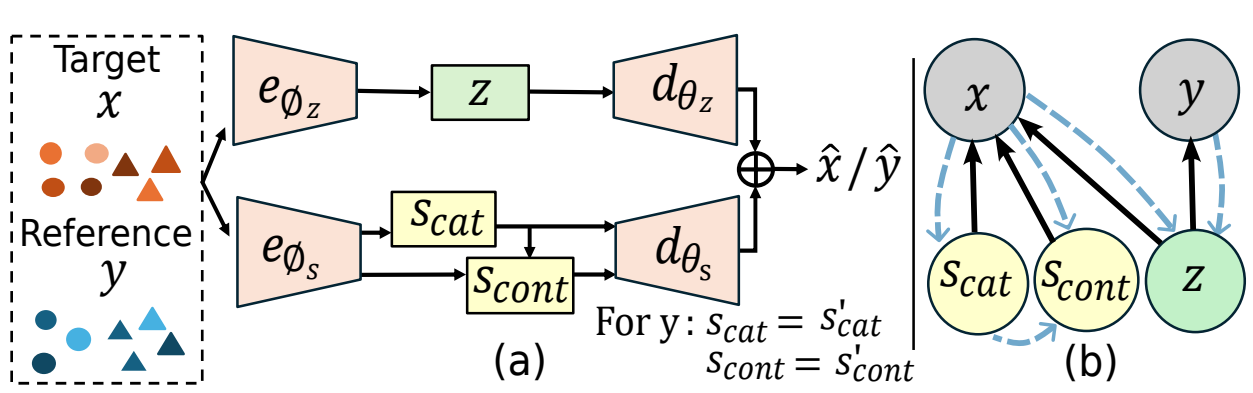

*Figure 1.* (a) CASL-VAE framework. Target ($x$) and reference ($y$) samples are encoded into a common latent $z$ and a hierarchical salient latent with discrete ($s_{\text{cat}}$) and continuous ($s_{\text{cont}}$) components. Decoders reconstruct $\hat{x}$ or $\hat{y}$ from these latent factors. For reference samples $y$, the salient latents are set to uninformative preset values $s'_{\text{cat}}$ and $s'_{\text{cont}}$. (b) Graphical model showing dependencies between observed data and latent variables. Solid arrows indicate generative paths; dashed arrows indicate inference.

### 3.2. Contrastive Latent Variables as Generative Factors

We introduce three latent variables: a continuous common latent variable $z \in \mathbb{R}^k$, a discrete (categorical) salient latent variable $s_{\text{cat}} \in \{1, ..., M\}$, and a continuous salient latent variable $s_{\text{cont}} \in \mathbb{R}^t$ conditioned on $s_{\text{cat}}$. The common latent models variation shared across both domains and is assumed to fully encode the variation present in the reference domain. The salient latent variables provide a **hierarchical representation** of target-specific variation, where the discrete component captures coarse subtype structure and the continuous component models finer-grained variation within these subtypes (Figure 1(b)).

So, the *prior* factorizes as

$$\begin{aligned} p(z, s_{\text{cat}}, s_{\text{cont}}) &= p(z)p(s_{\text{cat}}, s_{\text{cont}}) \\ &= p(z)\, p(s_{\text{cat}})\, p(s_{\text{cont}} \mid s_{\text{cat}}), \end{aligned} \tag{1}$$

where $p(z) = \mathcal{N}(0, I)$, $p(s_{\text{cat}}) = \text{Uniform}\{1, \ldots, M\}$, and $p(s_{\text{cont}} \mid s_{\text{cat}}) = \mathcal{N}(\mu_\psi(s_{\text{cat}}), I)$, where $\mu_\psi(s_{\text{cat}})$ is a learnable embedding function that maps each category $s_{\text{cat}}$ to a subtype-specific prior mean.

*Remark:* In neuroimaging applications, $z$ may encode demographic or scanner variability shared between healthy and diseased participants, while $s_{\text{cat}}$ and $s_{\text{cont}}$ may encode disease-specific pathological patterns of variability.

### 3.3. Optimizing the Paired Data Likelihood from Unpaired Data

Given $z$ and $s = (s_{\text{cat}}, s_{\text{cont}})$, the target and reference observations are independent and the conditional joint distribution factorizes as $p(x, y \mid z, s) = p(x \mid z, s)p(y \mid z)$. Similarly, the approximate posterior distribution factorizes as $q(z, s \mid x, y) = q(z \mid x, y)q(s \mid x)$. Using variational inference (Doersch, 2016), the variational lower bound on the joint log-likelihood is

$$\begin{aligned} \log p(x, y) \geq\ & \mathbb{E}_{q(z|x,y)q(s|x)} \log p(x|z, s) \\ & + \mathbb{E}_{q(z|x,y)} \log p(y|z) - KL(q(z|x, y)||p(z)) \\ & - KL(q(s|x)||p(s)). \end{aligned} \tag{2}$$

The derivation is provided in Appendix A.1. Importantly, this bound depends on $q(z|x, y)$, and therefore requires paired samples. However, if we impose posterior decoupling,

$$q(z|x, y) = q(z|x) = q(z|y), \tag{3}$$

the common latent becomes domain-invariant and can be inferred from either domain $x$ or $y$. This constraint can be implemented via regularization (Section 3.6). Under this assumption, Eq. 2 simplifies to a variational lower bound that can be optimized using unpaired samples:

$$\begin{aligned} \log p(x, y) \geq\ & \mathbb{E}_{q(z|x)q(s|x)} \log p(x|z, s) \\ & + \mathbb{E}_{q(z|y)} \log p(y|z) - \frac{1}{2} KL(q(z|x)||p(z)) \\ & - \frac{1}{2} KL(q(z|y)||p(z)) - KL(q(s|x)||p(s)). \end{aligned} \tag{4}$$

The KL term on the common latent is evenly split between domains to preserve symmetry and avoid over-regularizing the common latent. Therefore, under posterior decoupling, the joint variational lower bound decomposes into separate domain-specific terms corresponding to the target and reference marginals, yielding an objective equivalent in form to those optimized in Contrastive Analysis (Abid & Zou, 2019; Weinberger et al., 2022; Louiset et al., 2024). This provides a principled probabilistic justification for the common/salient latent decomposition, and explains why contrastive objectives can be viewed as implicitly maximizing a relaxed joint likelihood from unpaired data.

### 3.4. Parameterization of the Latent Inference Model and the Generative Model

The inference model consists of two neural networks: a common encoder network $e_{\phi_z}$ and a salient encoder network $e_{\phi_s}$, which together define a factorized variational posterior over the latent variables. For a target domain sample $x \in \mathcal{X}$, the variational *posterior* factorizes as

$$q_\phi(z, s_{\text{cat}}, s_{\text{cont}} \mid x) = q_{\phi_z}(z \mid x)\, q_{\phi_s}(s_{\text{cat}}, s_{\text{cont}} \mid x), \tag{5}$$

where $q_{\phi_s}(s_{\text{cat}}, s_{\text{cont}} \mid x) = q_{\phi_s}(s_{\text{cont}} \mid s_{\text{cat}}, x) q_{\phi_s}(s_{\text{cat}} \mid x)$.

The individual variational factors are parameterized as

$$q_{\phi_z}(z \mid x) = \mathcal{N}\big(z \mid \mu_z(x), \sigma_z^2(x) I\big)\,, \tag{6}$$

$$q_{\phi_s}(s_{\text{cat}} \mid x) = \text{Gumbel-Softmax}(\alpha(x), \tau), \tag{7}$$

$$q_{\phi_s}(s_{\text{cont}} \mid s_{\text{cat}},\ x) = \mathcal{N}(s_{\text{cont}} \mid \mu_s(s_{\text{cat}},\ x), I)\,. \tag{8}$$

Here, $e_{\phi_z}$ outputs $\mu_z(x)$ and $\sigma_z(x)$, while $e_{\phi_s}$ outputs the logits $\alpha(x)$ and the conditional mean $\mu_s(s_{\text{cat}}, x)$. The discrete latent is modeled using a Gumbel–Softmax distribution, where the temperature $\tau$ controls the relaxation of the categorical variable (Jang et al., 2017). During training, $s_{\text{cat}}$ lies on the probability simplex $\Delta^{M-1}$, enabling gradient-based optimization. At test time, it is discretized to a hard one-hot vector.

*Note*: For reference-domain samples $y \in \mathcal{Y}$, only $q_{\phi_z}(z \mid y)$ is inferred; the salient latents are fixed to an uninformative preset state $s' = (s'_{\text{cat}},\ s'_{\text{cont}})$ with $s'_{\text{cat}} = 1$ (reference category) and $s'_{\text{cont}} = 0$. This constraint prevents the salient latent from encoding reference-domain variation and encourages it to capture target-specific variability (Section 3.6).

For the generative model, we parameterize the relation between the latent variables and the observed data with two neural networks: a common decoder $d_{\theta_z}$ and a salient decoder $d_{\theta_s}$, giving

$$p_\theta(x \mid z, s_{\text{cat}}, s_{\text{cont}}) = \mathcal{N}\big(x \mid d_{\theta_z}(z) + d_{\theta_s}(s_{\text{cat}}, s_{\text{cont}}), I\big),$$
$$p_\theta(y \mid z) = \mathcal{N}\big(y \mid d_{\theta_z}(z) + d_{\theta_s}(s'_{\text{cat}}, s'_{\text{cont}}), I\big). \tag{9}$$

The additive decoder structure provides interpretability by explicitly separating common and salient contributions in the input space. For reference samples, the salient latent is fixed to $s' = (s'_{\text{cat}}, s'_{\text{cont}})$; thus, $d_{\theta_s}(s')$ acts as a constant baseline offset in the reference reconstruction.

This modeling allows direct interpretation and visualization of **prototypical target patterns of variation** as

$$\delta_{s_{\text{cat}}} = d_{\theta_s}(s_{\text{cat}}, \mu_\psi(s_{\text{cat}})) - d_{\theta_s}(s'_{\text{cat}} = 1, s'_{\text{cont}} = 0), \tag{10}$$

for each $s_{\text{cat}} \in \{2, \ldots, M\}$. This highlights the target-specific subtype patterns relative to the reference domain.

*Note*: We use a Gaussian likelihood for the observations because our application involves continuous brain volumes. However, the model can be adapted to other likelihoods (e.g., Bernoulli) for different data types.

### 3.5. Evidence Lower Bound (ELBO) for CASL-VAE

Expanding eq. 4 (see Appendix A.2 for details), the ELBO for CASL-VAE is

$$\mathcal{L}_{ELBO} = \mathcal{L}_x + \mathcal{L}_y, \tag{11}$$

where

$$\begin{aligned}\mathcal{L}_x &= \mathbb{E}_{q_{\phi_z}(z|x) q_{\phi_s}(s_{\text{cat}}, s_{\text{cont}}|x)} \log p(x \mid z, s_{\text{cat}}, s_{\text{cont}}) \\ &\quad - \frac{1}{2} KL(q_{\phi_z}(z \mid x) || p(z)) - KL(q_{\phi_s}(s_{\text{cat}} \mid x) || p(s_{\text{cat}})) \\ &\quad - \mathbb{E}_{q_{\phi_s}(s_{\text{cat}}|x)} [KL(q_{\phi_s}(s_{\text{cont}} \mid s_{\text{cat}}, x) || p(s_{\text{cont}} \mid s_{\text{cat}}))],\end{aligned} \tag{12}$$

$$\mathcal{L}_y = \mathbb{E}_{q_{\phi_z}(z|y)} \log p(y \mid z) - \frac{1}{2} KL(q_{\phi_z}(z \mid y) || p(z)). \tag{13}$$

This decomposition corresponds to the marginal ELBOs for the target and reference domains under the hierarchical latent structure.

### 3.6. Regularizations

To encourage a domain-invariant posterior $q_{\phi_z}(z \mid x) \approx q_{\phi_z}(z \mid y)$, we apply two complementary regularizers: a global distribution alignment term (MMD) and, under cross-domain transformations, a sample-level consistency term. Additionally, we introduce two more regularization terms to prevent the salient latent space from encoding reference-domain variation.

**Maximum Mean Discrepancy (MMD):** This term encourages the common latent distributions inferred from the two domains to be aligned (Gretton et al., 2012), Appendix A.3:

$$\mathcal{R}_{\text{MMD}} = \text{MMD}^2(q_{\phi_z}(z \mid x),\ q_{\phi_z}(z \mid y))\,. \tag{14}$$

**Sample-level Latent Consistency:** This term enforces that the common latent representation is stable under cross-domain transformation from $\mathcal{X}$ to $\mathcal{Y}$, which provides sample-level alignment $q_{\phi_z}(z \mid x) \approx q_{\phi_z}(z \mid \tilde{y})$:

$$\mathcal{R}_{\text{LC}} = \mathbb{E}_{x \sim p(x),\ z \sim q_{\phi_z}(z|x)} \left\| \mu_z(\tilde{y}) - \mu_z(x) \right\|_1, \tag{15}$$

$$\text{where} \quad \tilde{y} = d_{\theta_z}(z) + d_{\theta_s}(s'_{\text{cat}}, s'_{\text{cont}}). \tag{16}$$

Here, $\tilde{y}$ is a conditionally generated paired reference sample obtained by decoding $z$ with the salient latent fixed to the uninformative preset values.

**Semi-supervised Binary Cross-Entropy Loss:** To prevent the discrete salient latent from capturing reference-domain structure, we use domain labels (reference vs. target) as weak supervision. Reference samples are encouraged to map to category $s'_{\text{cat}} = 1$, while target samples are discouraged from this category, leaving the remaining $M-1$ categories for unsupervised target subtype discovery:

$$\mathcal{R}_{\text{BCE}} = -\mathbb{E}_{y\sim p(y)} \log q_{\phi_s}(s_{\text{cat}} = 1 \mid y) - \mathbb{E}_{x\sim p(x)} \log\big(1 - q_{\phi_s}(s_{\text{cat}} = 1 \mid x)\big). \quad (17)$$

**Reference Regularization of the Continuous Salient Latent:** Although $\mathcal{R}_{\text{BCE}}$ separates reference and target samples in the discrete latent space, the continuous latent $s_{\text{cont}}$ could still encode reference-domain variation. To prevent this, we regularize both the posterior and learned prior of $s_{\text{cont}}$ for reference samples:

$$\begin{aligned}\mathcal{R}_{\text{ref}} &= \mathrm{KL}(q_{\phi_s}(s_{\text{cont}} \mid y, s'_{\text{cat}}) \,\|\, \mathcal{N}(0, I)) \\ &\quad + \mathrm{KL}(p(s_{\text{cont}} \mid s'_{\text{cat}}) \,\|\, \mathcal{N}(0, I))\,, \\ &= \frac{1}{2}\big\|\mu_s(y, s'_{\text{cat}})\big\|_2^2 + \frac{1}{2}\big\|\mu_\psi(s'_{\text{cat}})\big\|_2^2. \end{aligned} \quad (18)$$

where the final expression follows from both distributions having identity covariance. This penalty drives both posterior and prior means toward zero, making $s_{\text{cont}}$ uninformative for reference samples.

*Information-Theoretic Interpretation:* $\mathcal{R}_{\text{BCE}}$ and $\mathcal{R}_{\text{ref}}$ jointly minimize an upper bound on the mutual information between reference domain $\mathcal{Y}$ and the salient latent $(s_{\text{cat}}, s_{\text{cont}})$ (see Appendix A.4). The former collapses the discrete salient latent for $\mathcal{Y}$, while the latter suppresses information in the continuous latent, ensuring that the salient latent variables capture only target-specific variation.

### 3.7. Overall Training Objective

The final training objective minimizes the negative ELBO along with the additional regularizers:

$$\mathcal{J} = -\mathcal{L}_{ELBO} + \lambda\,\mathcal{R}_{\text{MMD}} + \eta\,\mathcal{R}_{\text{LC}} + \zeta\,\mathcal{R}_{\text{BCE}} + \gamma\,\mathcal{R}_{\text{ref}}, \quad (19)$$

where $\lambda$, $\eta$, $\zeta$, and $\gamma$ control the strength of each regularizer. All expectations are approximated via Monte Carlo sampling with the reparameterization trick.

### 3.8. Inference and Generation

Besides providing a rich latent representation of reference-target data, the trained model supports three inference/generation modes:

1. **Subtype inference:** For each sample, the salient encoder outputs logits $\alpha(\cdot)$ over the $M$ categories. After softmax, the sample is assigned to the category with the highest probability. Category 1 corresponds to the reference domain, while the remaining $M-1$ categories represent target subtypes.

2. **Conditional paired-sample generation:** Given a real sample, the common encoder infers $z \sim q_{\phi_z}(z \mid \cdot)$. Decoding $z$ with the preset salient state $s' = (s'_{\text{cat}}, s'_{\text{cont}})$ yields a paired reference-domain sample $\tilde{y}$, while decoding with alternative salient values yields paired target-domain samples $\tilde{x}$.

3. **Synthetic data generation:** Paired samples share the same common latent $z$, while unpaired samples have independent $z$. To generate synthetic paired samples, first draw $z \sim p(z)$, then decode it with the preset state $s'$ to obtain a reference sample, and with varying salient values $(s_{\text{cat}}, s_{\text{cont}})$ to obtain corresponding target samples. To generate unpaired samples, draw separate $z$ values for each domain, and repeat.

## 4. Experiments

Our experiments on semi-synthetic and real data assess CASL-VAE's performance in (i) target–subtype discovery, (ii) paired-sample generation, and (iii) real-world clinical utility. For all experiments, we set the common latent dimension to $k = 15$, the discrete salient latent to $M$ categories (selected via a cross-validation procedure; see Section 4.2), and the continuous salient latent to $t = 2$ dimensions, enabling direct visualization of within-subtype variation in 2-D.

All additional implementation and training details are provided in Appendix A.5.

**Data Preparation:** For all experiments, we used T1-weighted Magnetic Resonance Imaging (T1w MRI) and clinical measurements from the iSTAGING consortium (Habes et al., 2021), which integrates data from multiple aging and neurodegenerative disease cohorts. From baseline T1w MRI scans, brain volumes of 145 regions of interest (ROIs) were extracted using a multi-atlas label fusion method (Doshi et al., 2016), harmonized to remove site effects (Pomponio et al., 2020), and used as input features.

Across all experiments, unpaired reference–target samples were used for model training. For each train/test split, all samples were residualized to remove covariate effects of age, sex, and intracranial volume (ICV) using linear regression coefficients estimated from the *training reference* data. Finally, the adjusted features were standardized with respect to the training reference distribution.

*Table 1.* ARI clustering performance for semi-synthetic data under mild (10–20%) and typical (10–30%) atrophy rates. Values show mean and 95% confidence interval (CI) over 10 repeated cross-validation analysis. ARI ranges from 0 (random clustering) to 1 (perfect clustering). **Bold**: best.

| Model | ARI ↑ (target-only) | | ARI ↑ (target+reference) | |
|---|---|---|---|---|
| | 10–20% | 10–30% | 10–20% | 10–30% |
| GMM | 0.001 (−0.005, 0.007) | 0.232 (0.070, 0.394) | 0.001 (−0.006, 0.008) | 0.043 (0.009, 0.076) |
| GMVAE | 0.038 (0.006, 0.069) | 0.284 (0.129, 0.439) | – | – |
| K-means | −0.002 (−0.009, 0.004) | 0.533 (0.423, 0.643) | 0.006 (0.001, 0.012) | 0.112 (0.093, 0.130) |
| cVAE | 0.183 (0.124, 0.242) | 0.654 (0.581, 0.726) | 0.009 (0.000, 0.019) | 0.041 (0.018, 0.064) |
| SepVAE | 0.251 (0.196, 0.306) | 0.611 (0.558, 0.664) | 0.159 (0.130, 0.188) | 0.451 (0.380, 0.523) |
| Smile-GAN | 0.183 (0.097, 0.293) | **0.657 (0.508, 0.750)** | – | – |
| **CASL-VAE** | **0.468 (0.371, 0.564)** | 0.620 (0.535, 0.706) * | **0.587 (0.521, 0.653)** | **0.697 (0.672, 0.722)** |

**Note:** * ARI (target-only) of CASL-VAE is not significantly different from best (CIs overlap).

### 4.1. Semi-synthetic Data Experiments

To have ground-truth disease patterns and severity, we simulated disease effects in real cognitively normal (CN) samples. Specifically, we randomly split 988 CN (age 45-79) participants into a CN reference ($N = 494$) group and a pseudo-target ($N = 494$) group. The pseudo-target group was then randomly divided into three disease subtypes (S1–S3, Appendix Table 5), each defined by a unique pattern of 14 ROIs (with some overlap across subtypes). Consequently, the discrete salient latent $s_{\text{cat}}$ has dimension $M = 4$ (one reference category and three pseudo-target subtypes). For each pseudo-target sample, subtype-specific ROIs were reduced by a rate uniformly sampled between 10–30%, representing the typical atrophy range. To evaluate model robustness to effect size, additional datasets were created with narrower ranges: 0–10% (extremely mild), 10–20% (mild), and 20–30% (moderate). Additionally, to evaluate paired sample generation, we applied the above simulation to held-out CN samples ($N = 496$) to generate corresponding *paired* pseudo-target samples, creating a semi-synthetic test set with ground-truth pairings.

**Evaluation metrics:** Subtype discovery is evaluated using Adjusted Rand Index (ARI) (Hubert & Arabie, 1985). Reconstruction quality is measured using mean absolute error (MAE) between input and reconstruction output. For paired sample generation, cosine similarity is used to measure similarity between generated and ground-truth paired reference samples.

**Baseline model comparisons:** We compare CASL-VAE to several baselines: (i) standard clustering (K-means, Gaussian mixture modeling (GMM)), (ii) deep clustering (GMVAE (Shu et al., 2017)), (iii) semi-supervised clustering (Smile-GAN (Yang et al., 2021)), (iv) standard generative modeling (cVAE (Doersch, 2016)), and (v) generative model for contrastive analysis (SepVAE (Louiset et al., 2024)). To evaluate clustering performance of cVAE and SepVAE, we apply K-means to the posterior means of their latent representations (unified latent space for cVAE; salient latent space for SepVAE). We detail the experiment procedure in Appendix B.2.

### 4.2. Real Data Experiment

We applied CASL-VAE to the ADNI dataset (Beckett et al., 2015), which includes cognitively normal (CN) individuals as well as participants with mild cognitive impairment (MCI) or Alzheimer's disease (AD). CN participants served as the reference domain, while MCI and AD participants formed the target domain. The model was trained on ADNI2/GO participants (206 reference samples, 660 target samples) and evaluated on held-out ADNI1 participants (227 reference samples, 588 target samples). Details on available clinical variables used for downstream analysis can be found in Appendix B.5.

**Selecting the optimal dimension of $s_{\text{cat}}$ ($M$).** In real data, the true number of target subtypes is unknown. To select $M$, we assessed clustering stability across repeated hold-out cross-validation. For each candidate $M \in \{2, 3, 4, 5, 6\}$, we performed 10 repetitions of hold-out CV: in each repetition the model was trained on 80% of the training data, and subtype assignments were obtained for the full training set. We computed the ARI between all pairs of clusterings across the 10 repetitions, and used the mean pairwise ARI as the stability score for each $M$. The value of $M$ with the highest stability score was selected.

### 4.3. Results on Semi-synthetic Data

**Subtype discovery.** In addition to target-subtype discovery (target-only clustering), we evaluate whether reference samples are correctly separated from target subtypes in a joint (target + reference) clustering setting. Table 1 reports clustering performance under two semi-synthetically modeled atrophy regimes: a typical regime (10–30%) and a mild regime (10–20%). While the typical regime corresponds to standard simulated effect sizes, we additionally report

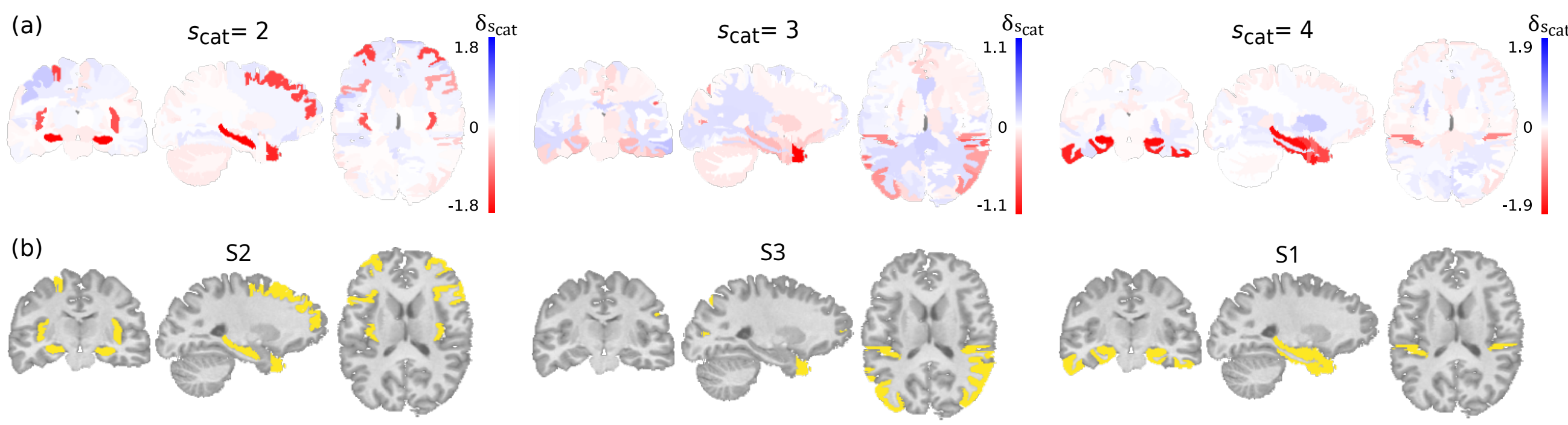

*Figure 2.* Results from Semi-synthetic experiment with 10–30% atrophy. a) Prototypical target patterns $\delta_{s_{\text{cat}}}$ learned by CASL-VAE for $s_{\text{cat}} = 2, 3, 4$. Negative values (red) indicate volumetric reductions with respect to reference. b) Ground-truth subtype (S2,S3,S1) patterns; yellow = binary mask.

results in the mild regime to highlight a more challenging setting with weaker effect sizes.

Under the typical atrophy setting, CASL-VAE performs on par with the best baseline (Smile-GAN), suggesting that pronounced disease effects enable multiple methods to recover subtype structure. Importantly, CASL-VAE provides an interpretable representation of the learned salient latent structure. Figure 2 shows brain maps of the prototypical target patterns learned by the model alongside the ground-truth subtype masks used in simulation. We observe near-perfect correspondence between the inferred patterns and the simulated masks, providing visual evidence that CASL-VAE recovers meaningful and interpretable subtype factors. In addition to recovering subtype-level patterns, CASL-VAE also captures within-subtype variability through the continuous salient latent space, as shown in Figure 3.

In the mild atrophy setting, CASL-VAE significantly outperforms all baselines in target-only clustering. It also achieves the highest ARI in the joint target + reference setting in both the typical and mild settings, demonstrating that it can simultaneously discover target subtypes and distinguish them from the reference domain. These gains can be attributed to CASL-VAE's hierarchical salient latent structure. Importantly, this structured latent factorization does not compromise its reconstruction quality (Appendix Tables 6 and 7).

*Table 2.* Cosine similarity (mean, 95% CI) between generated and ground-truth paired reference samples in semi-synthetic data under mild (10–20%) and typical (10–30%) atrophy. **Bold**: best.

| Model | 10–20% | 10–30% |
|---|---|---|
| cVAE | 0.5471 (0.5428, 0.5514) | 0.5208 (0.5170, 0.5247) |
| SepVAE | 0.5287 (0.5188, 0.5387) | 0.5409 (0.5326, 0.5493) |
| **CASL-VAE** | **0.5835 (0.5777, 0.5894)** | **0.5816 (0.5747, 0.5885)** |

We further performed **ablation studies** to assess sensitivity to target-specific effect size and robustness to regularization hyperparameters, with detailed results reported in Appendix B.3.

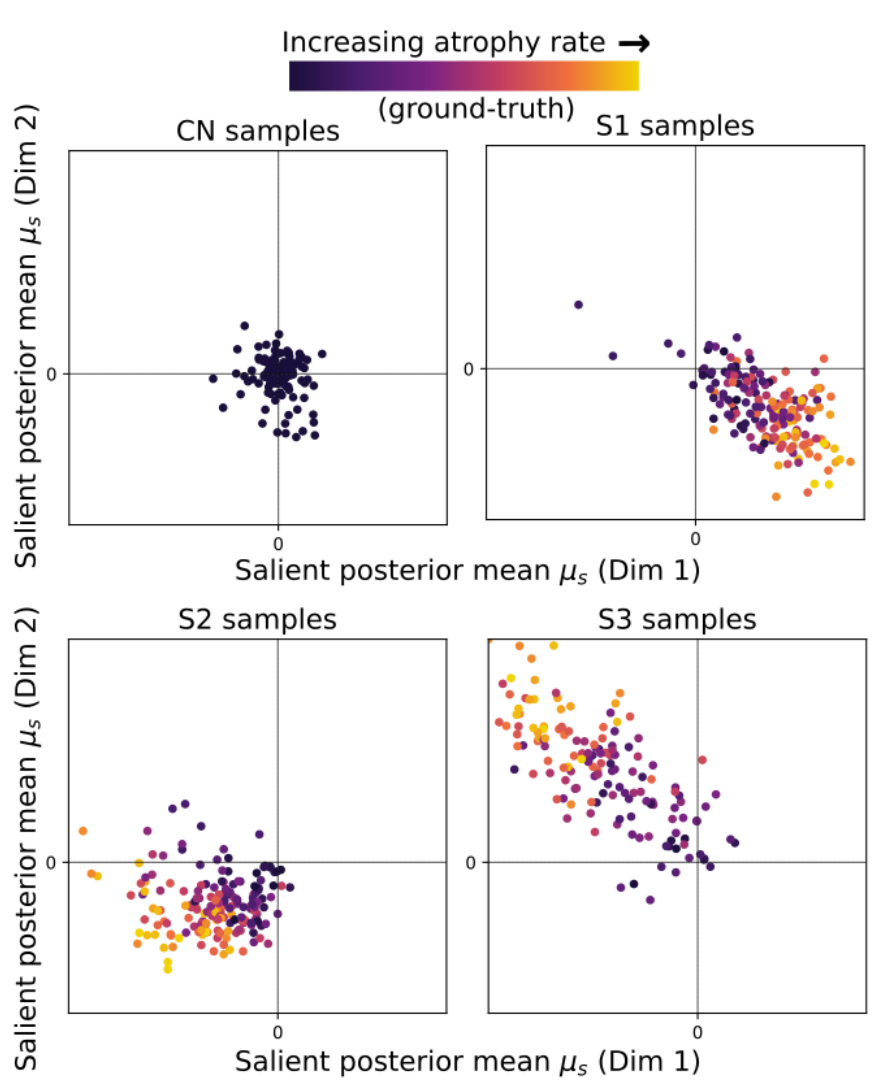

*Figure 3.* Conditional (continuous) salient 2-D latent means from held-out CN and pseudo-target samples (simulated with 10-30% atrophy). Each point represents a sample's posterior mean $\mu_s(\cdot)$. The cross-hair marks the preset value $s'_{\text{cont}}$. Color indicates the ground-truth simulated atrophy rate.

**Conditional paired-sample generation.** We evaluated paired sample generation on semi-synthetic test datasets with paired reference and target samples. For each target sample, we generated a paired reference $\tilde{y}$ (Eq. 16) and measured its cosine similarity with the ground-truth paired reference sample, assessing their semantic similarity. Across both mild and typical atrophy rate, CASL-VAE achieves the highest cosine similarity compared to baseline generative models (Table 2), indicating better semantic consistency of the generated paired samples.

### 4.4. Results on Real Disease Data

On the ADNI dataset, CASL-VAE achieves the highest clustering stability when the discrete salient latent has dimension $M$=3 (Appendix Figure 6). Figure 4 illustrates the resulting prototypical target-subtype patterns learned by this model.

The first pattern (top, $\delta_{s_{\text{cat}}=2}$) exhibits diffuse brain atrophy, including in regions like the frontal lobe, entorhinal cortex, and angular gyrus, and appears milder in comparison to the second pattern. The second pattern (bottom, $\delta_{s_{\text{cat}}=3}$) shows more pronounced volumetric reductions concentrated in the medial temporal lobe (e.g. temporal gyrus and hippocampus).

We further validated the clinical utility of CASL-VAE by measuring the mean absolute target-specific salient variation inferred in each participant in the held-out ADNI1 dataset. This serves as a deviation index, quantifying each participant's overall deviation relative to the reference domain (Appendix B.5). This index showed moderate to strong associations with clinical and biomarker variables for AD (Appendix Table 8), suggesting that the inferred target-specific variation is clinically relevant.

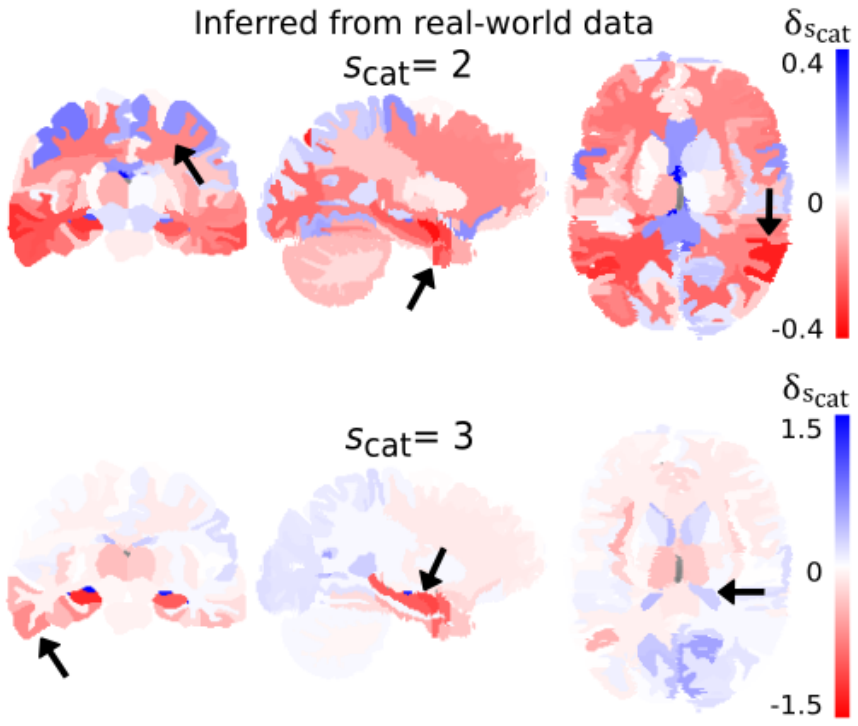

*Figure 4.* Prototypical target patterns $\delta_{s_{\text{cat}}}$ learned in ADNI2/GO MCI and AD participants relative to CN participants; arrows point to some clinically relevant regions. Negative values (red) indicate volumetric reductions with respect to CN group.

## 5. Conclusion

We introduced CASL-VAE, a contrastive generative framework for learning structured latent representations from unpaired reference–target data. The model provides a data-driven, interpretable approach for uncovering how target populations deviate from the reference.

By factorizing latent structure into a shared continuous space and a hierarchical salient space that captures discrete subtypes and continuous within-subtype variation, CASL-VAE enables unsupervised target-subtype discovery, interpretable cross-domain comparisons, and paired-sample generation.

In semi-synthetic neuroimaging experiments, CASL-VAE outperformed clustering, generative modeling, and contrastive analysis baselines, particularly in regimes with subtle effects where heterogeneity and entanglement challenge standard methods. In real Alzheimer's disease data, it uncovered biologically plausible patterns of brain variation (Coupé et al., 2019), highlighting its potential for exploratory analysis of disease heterogeneity.

More broadly, CASL-VAE offers a probabilistically grounded perspective on contrastive analysis, showing how structured latent inductive biases can bridge disentanglement, clustering, and generative reasoning in unpaired-data settings. Future work may extend the framework to longitudinal data, richer hierarchical structures, or alternative likelihoods, and explore applications in other domains requiring analysis of reference–target variation.

## Impact Statement

CASL-VAE is a generative framework that has the potential to support exploratory data analysis in domains where paired observations are difficult or impossible to obtain. In biomedical research, the framework may facilitate the study of disease heterogeneity and subgroup structure, particularly in high-dimensional biological data, supporting hypothesis generation and exploratory analysis. More broadly, the framework is applicable to contrastive population analysis in domains such as genomics, epidemiology, and social sciences.

The model is intended for research use rather than clinical decision-making. As with other representation-learning methods, biases in the training data may be reflected in the learned latent structure, and improper use without rigorous validation could lead to misleading conclusions.

# A. Appendix: Methods

### A.1. Derivation of the ELBO on the joint log-likelihood

In the main text we defined the joint generative model $p(x, y, z, s) = p(z)p(s)p(x \mid z, s)p(y \mid z)$, and the variational posterior $q(z, s \mid x, y) = q(z \mid x, y)q(s \mid x)$. To define the joint log-likelihood, we introduce the variational posterior $q(z, s \mid x, y)$ as:

$$\log p(x,y) = \log \int p(x,y,z,s)\,dz\,ds = \log \int q(z,s \mid x,y)\frac{p(x,y,z,s)}{q(z,s \mid x,y)}\,dz\,ds = \log \mathbb{E}_{q(z,s|x,y)}\left[\frac{p(x,y,z,s)}{q(z,s \mid x,y)}\right]. \quad (20)$$

Applying Jensen's inequality yields

$$\log p(x,y) \geq \mathbb{E}_{q(z,s|x,y)}\left[\log \frac{p(x,y,z,s)}{q(z,s \mid x,y)}\right]. \quad (21)$$

Using the conditional independence $p(x, y \mid z, s) = p(x \mid z, s)p(y \mid z)$ and the posterior factorization $q(z, s \mid x, y) = q(z \mid x, y)q(s \mid x)$, we obtain

$$\log p(x,y) \geq \mathbb{E}_{q(z|x,y)q(s|x)}[\log p(x \mid z,s)] + \mathbb{E}_{q(z|x,y)}[\log p(y \mid z)] - \mathrm{KL}(q(z \mid x,y)\|p(z)) - \mathrm{KL}(q(s \mid x)\|p(s)),$$

which is the variational lower bound of the joint log marginal likelihood, as given in Eq. (2) in the main text.

Assuming posterior decoupling $q(z \mid x, y) = q(z \mid x) = q(z \mid y)$[2], the KL term on the common latent can be split as:

$$\mathrm{KL}(q(z \mid x,y)\|p(z)) = \frac{1}{2}\mathrm{KL}(q(z \mid x)\|p(z)) + \frac{1}{2}\mathrm{KL}(q(z \mid y)\|p(z)) \quad (22)$$

Thus, Eq. (2) can be rewritten as Eq. (4).

### A.2. Derivation of the ELBO for CASL-VAE

Starting from the ELBO in Eq. (4),

$$\begin{aligned}\log p(x,y) \geq{}& \mathbb{E}_{q(z|x)q(s|x)}[\log p(x \mid z,s)] + \mathbb{E}_{q(z|y)}[\log p(y \mid z)] - \tfrac{1}{2}\mathrm{KL}(q(z \mid x)\|p(z)) - \tfrac{1}{2}\mathrm{KL}(q(z \mid y)\|p(z)) \\ &- \mathrm{KL}(q(s \mid x)\|p(s)),\end{aligned} \quad (23)$$

The salient latent variable is hierarchical, $s = (s_{\text{cat}}, s_{\text{cont}})$, with posterior factorization $q(s \mid x) = q(s_{\text{cat}} \mid x)\, q(s_{\text{cont}} \mid s_{\text{cat}}, x)$, and prior factorization $p(s) = p(s_{\text{cat}})\, p(s_{\text{cont}} \mid s_{\text{cat}})$.

The KL divergence for the salient latent is expanded as

$$\begin{aligned}\mathrm{KL}(q(s \mid x)\|p(s)) &= \mathbb{E}_{q(s|x)}\left[\log \frac{q(s \mid x)}{p(s)}\right] \\ &= \mathbb{E}_{q(s_{\text{cat}}|x)q(s_{\text{cont}}|s_{\text{cat}},x)}\left[\log \frac{q(s_{\text{cat}} \mid x)q(s_{\text{cont}} \mid s_{\text{cat}},x)}{p(s_{\text{cat}})p(s_{\text{cont}} \mid s_{\text{cat}})}\right] \\ &= \mathbb{E}_{q(s_{\text{cat}}|x)}\left[\log \frac{q(s_{\text{cat}} \mid x)}{p(s_{\text{cat}})}\right] + \mathbb{E}_{q(s_{\text{cat}}|x)}\left[\mathbb{E}_{q(s_{\text{cont}}|s_{\text{cat}},x)} \log \frac{q(s_{\text{cont}} \mid s_{\text{cat}},x)}{p(s_{\text{cont}} \mid s_{\text{cat}})}\right] \\ &= \mathrm{KL}(q(s_{\text{cat}} \mid x)\|p(s_{\text{cat}})) + \mathbb{E}_{q(s_{\text{cat}}|x)}\left[\mathrm{KL}(q(s_{\text{cont}} \mid s_{\text{cat}},x)\|p(s_{\text{cont}} \mid s_{\text{cat}}))\right].\end{aligned} \quad (24)$$

Substituting Eq. (24) into Eq. (23) gives

$$\begin{aligned}\log p(x,y) \geq{}& \mathbb{E}_{q(z|x)q(s_{\text{cat}},\, s_{\text{cont}}|x)}[\log p(x \mid z, s_{\text{cat}}, s_{\text{cont}})] + \mathbb{E}_{q(z|y)}[\log p(y \mid z)] - \tfrac{1}{2}\mathrm{KL}(q(z \mid x)\|p(z)) \\ &- \tfrac{1}{2}\mathrm{KL}(q(z \mid y)\|p(z)) - \mathrm{KL}(q(s_{\text{cat}} \mid x)\|p(s_{\text{cat}})) - \mathbb{E}_{q(s_{\text{cat}}|x)}\left[\mathrm{KL}(q(s_{\text{cont}} \mid s_{\text{cat}},x)\|p(s_{\text{cont}} \mid s_{\text{cat}}))\right].\end{aligned} \quad (25)$$

[2]Posterior decoupling is enforced via regularization using $R_{MMD}$ and $R_{LC}$ as described in Section 3.6, which encourages $q(z \mid x) \approx q(z \mid y)$.

Finally, separating the domain-specific terms as

$$\mathcal{L}_x = \mathbb{E}_{q(z|x)q(s_{\text{cat}},\, s_{\text{cont}}|x)}[\log p(x \mid z, s_{\text{cat}}, s_{\text{cont}})] - \tfrac{1}{2}\text{KL}(q(z \mid x)\|p(z)) \\ - \text{KL}(q(s_{\text{cat}} \mid x)\|p(s_{\text{cat}})) - \mathbb{E}_{q(s_{\text{cat}}|x)}\left[\text{KL}(q(s_{\text{cont}} \mid s_{\text{cat}}, x)\|p(s_{\text{cont}} \mid s_{\text{cat}}))\right], \tag{26}$$

$$\mathcal{L}_y = \mathbb{E}_{q(z|y)}[\log p(y \mid z)] - \tfrac{1}{2}\text{KL}(q(z \mid y)\|p(z)), \tag{27}$$

we obtain the ELBO for CASL-VAE, as defined in Eq. (11) (where it is written in terms of parameterized distributions):

$$\mathcal{L}_{ELBO} = \mathcal{L}_x + \mathcal{L}_y.$$

### A.3. MMD Regularization for Posterior Decoupling

We use maximum mean discrepancy (MMD) as one of the regularizations to align the common latent distributions $q(z|x)$ and $q(z|y)$ (Section 3.6):

$$\mathcal{R}_{\text{MMD}} = \text{MMD}^2(q(z|x), q(z|y)) = \mathbb{E}_{z_x, z_x'}[k(z_x, z_x')] + \mathbb{E}_{z_y, z_y'}[k(z_y, z_y')] - 2\mathbb{E}_{z_x, z_y}[k(z_x, z_y)]$$

where $k(\cdot,\cdot)$ is a sum of RBF kernels: $k(a,b) = \sum_{\gamma\in\Gamma}\exp(-\gamma\|a-b\|^2)$. We use $\Gamma = \{10^{-3}, 10^{-2}, 10^{-1}, 10^{0}, 10^{1}, 10^{2}\}$, corresponding to $\gamma = \frac{1}{2\sigma^2}$

### A.4. Information-Theoretic View of Salient Latent Regularization for the Reference Domain

We analyze how the reference-domain regularizers $\mathcal{R}_{\text{BCE}}$ and $\mathcal{R}_{\text{ref}}$ reduce the mutual information between reference samples $\mathcal{Y}$ and the salient latent $(s_{\text{cat}}, s_{\text{cont}})$.

**Mutual information and variational bound:** For random variables $X$ and $Z$, the mutual information is

$$I(X;Z) = H(Z) - H(Z \mid X), \tag{28}$$

where $H(\cdot)$ denotes Shannon entropy. When $Z$ is represented by any choice of variational posterior $q(z \mid x)$ and prior $p(z)$, a standard variational upper bound (Hoffman & Johnson, 2016) is

$$I(X;Z) \;\leq\; \mathbb{E}_{p(x)}[D_{\text{KL}}(q(z \mid x) \,\|\, p(z))]\,. \tag{29}$$

**Mutual Information between reference domain and hierarchical salient latent.** For reference samples $y \in \mathcal{Y}$, the hierarchical salient latent posterior and prior decomposes as

$$q(s_{\text{cat}}, s_{\text{cont}} \mid y) = q(s_{\text{cat}} \mid y)\, q(s_{\text{cont}} \mid s_{\text{cat}}, y), \qquad p(s_{\text{cat}}, s_{\text{cont}}) = p(s_{\text{cat}})\, p(s_{\text{cont}} \mid s_{\text{cat}}).$$

By the chain rule,

$$I(Y; s_{\text{cat}}, s_{\text{cont}}) = I(Y; s_{\text{cat}}) + I(Y; s_{\text{cont}} \mid s_{\text{cat}}). \tag{30}$$

For the continuous part, applying the variational bound yields

$$I(Y; s_{\text{cont}} \mid s_{\text{cat}}) \leq \mathbb{E}_{p(y)\, q(s_{\text{cat}}|y)}\left[D_{\text{KL}}\left(q(s_{\text{cont}} \mid s_{\text{cat}}, y) \,\|\, p(s_{\text{cont}} \mid s_{\text{cat}})\right)\right]. \tag{31}$$

Thus,

$$I(Y; s_{\text{cat}}, s_{\text{cont}}) \leq I(Y; s_{\text{cat}}) + \mathbb{E}_{p(y)\, q(s_{\text{cat}}|y)}\left[D_{\text{KL}}(q(s_{\text{cont}} \mid s_{\text{cat}}, y) \,\|\, p(s_{\text{cont}} \mid s_{\text{cat}}))\right], \tag{32}$$

which quantifies an upper bound on the information encoded in the salient latent variables about the reference samples.

**Effect of regularization terms $\mathcal{R}_{BCE}$ and $\mathcal{R}_{ref}$:** For reference domain $\mathcal{Y}$,

**1.** Minimizing $\mathcal{R}_{BCE}$ pushes $q(s_{\text{cat}} \mid y)$ toward a point mass at the preset reference category ($s'_{\text{cat}} = 1$). This reduces the conditional entropy $H(s_{\text{cat}} \mid Y)$ and the marginal entropy $H(s_{\text{cat}})$, thereby minimizing $I(Y; s_{\text{cat}})$.

**2.** The KL-divergence regularization $\mathcal{R}_{\text{ref}}$ on continuous salient latent enforces that both the posterior $q(s_{\text{cont}} \mid s_{\text{cat}} = 1, y)$ and the learned prior $p(s_{\text{cont}} \mid s_{\text{cat}} = 1)$ match $\mathcal{N}(0, I)$. By the variational upper bound in Eq. 31, minimizing $\mathcal{R}_{\text{ref}}$ thereby minimizes the upper bound on $I(Y; s_{\text{cont}} \mid s_{\text{cat}})$.

Therefore, $\mathcal{R}_{\text{BCE}}$ and $\mathcal{R}_{\text{ref}}$ jointly minimize an upper bound on $I(Y; s_{\text{cat}}, s_{\text{cont}})$ (Eq. 30), making the salient latent uninformative about the reference domain and forcing it to encode target-specific variation.

## A.5. Implementation Details

### A.5.1. Model Architecture

The overall architecture of CASL-VAE is illustrated in Fig. 1a. The common encoder $e_{\phi_z}$ maps input data to the parameters of the common latent posterior, producing a mean vector and a log-standard-deviation vector (each of dimension 15). The salient encoder $e_{\phi_s}$ first projects the input into a 36-dimensional embedding. This embedding is then mapped to $M$-dimensional logits, which parameterize the discrete salient posterior $q_{\phi_s}(s_{\text{cat}} \mid x)$. The sampled discrete salient variable is concatenated with the 36-dimensional embedding and passed through additional layers to produce the conditional continuous salient posterior mean (dimension 2), corresponding to $q_{\phi_s}(s_{\text{cont}} \mid s_{\text{cat}}, x)$.

The common decoder $d_{\theta_z}$ reconstructs the shared component from the sampled common latent representation, while the salient decoder $d_{\theta_s}$ reconstructs the target-specific component from the concatenated sampled discrete and continuous salient latents. The outputs of the two decoders are added to form the final reconstruction of the input (dimension 145).

Table 3 summarizes the network architectures of the common encoder and decoder $e_{\phi_z}$ and $d_{\theta_z}$. Table 4 summarizes the network architectures of the salient encoder and decoder $e_{\phi_s}$ and $d_{\theta_s}$, as well as the learned prior mean embedding $\mu_\psi$.

*Table 3.* Architecture of common encoder network $e_{\phi_z}$ and decoder network $d_{\theta_z}$

| Layer | Input Size | Bias Term | Leaky ReLU $\alpha$ | Output Size |
|---|---|---|---|---|
| **Common Encoder** $e_{\phi_z}$ | | | | |
| Linear1 + Leaky-ReLU | 145×1 | Yes | 0.2 | 72×1 |
| Linear2 + Leaky-ReLU | 72×1 | Yes | 0.2 | 36×1 |
| Linear3 | 36×1 | Yes | NA | 30×1 |
| *($e_{\phi_z}$ output split into 15-dim mean $\mu_z$ and log-standard-deviation $\log \sigma_z$)* | | | | |
| **Common Decoder** $d_{\phi_z}$ | | | | |
| Linear1 + Leaky-ReLU | 15×1 | Yes | 0.2 | 36×1 |
| Linear2 + Leaky-ReLU | 36×1 | Yes | 0.2 | 72×1 |
| Linear3 | 72×1 | Yes | NA | 145×1 |

*Table 4.* Architecture of salient encoder network $e_{\phi_s}$, decoder network $d_{\theta_s}$, and the learned prior mean embedding network $\mu_\psi$.

| Layer | Input Size | Bias Term | Leaky ReLU $\alpha$ | Output Size |
|---|---|---|---|---|
| **Salient Encoder** $e_{\phi_s}$ | | | | |
| **Input embedding** $e_{\phi_s}^{\text{embed}}$ | | | | |
| Linear1 + Leaky-ReLU | 145×1 | Yes | 0.2 | 72×1 |
| Linear2 + Leaky-ReLU | 72×1 | Yes | 0.2 | 36×1 |
| **Discrete salient component** $e_{\phi_s}^{\text{cat}}$ | | | | |
| Linear3 + Leaky-ReLU | 36×1 | Yes | 0.2 | 36×1 |
| Linear4 | 36×1 | Yes | NA | $M$ ×1 |
| **Continuous salient component** $e_{\phi_s}^{\text{cont}}$ | | | | |
| Linear5 + Leaky-ReLU | 36×1 | Yes | 0.2 | 36×1 |
| Linear6 | 36×1 | Yes | NA | 2×1 |
| *($e_{\phi_s}^{cat}$ outputs $M$-dim logits, $e_{\phi_s}^{cont}$ outputs 2-dim mean $\mu_s$ )* | | | | |
| **Salient prior mean embedding** $\mu_\psi$ | | | | |
| Linear1 + Leaky-ReLU | $M$×1 | Yes | 0.2 | 36×1 |
| Linear2 | 36×1 | Yes | NA | 2×1 |
| **Salient Decoder** $d_{\phi_s}$ | | | | |
| **Projection of discrete salient latent** | | | | |
| Linear1 | $M$×1 | Yes | NA | 9×1 |
| **Projection of continuous salient latent** | | | | |
| Linear2 | 2×1 | Yes | NA | 9×1 |
| **Concatenate discrete and continuous salient latent and decode** | | | | |
| Linear3 + Leaky-ReLU | 18×1 | Yes | 0.2 | 36×1 |
| Linear4 + Leaky-ReLU | 36×1 | Yes | 0.2 | 72×1 |
| Linear5 | 72×1 | Yes | NA | 145×1 |

### A.5.2. Training Details

The model was trained using the ADAM optimizer (Kingma & Ba, 2015) with a learning rate of $1 \times 10^{-3}$. The optimizer parameters were set to $\beta_1 = 0.5$ and $\beta_2 = 0.999$. For all experiments, we used the following regularization hyperparameters: $\lambda = 100$, $\eta = 10$, $\zeta = 1$, and $\gamma = 1$. We found the model to be largely robust to variations in these hyperparameters, as demonstrated in Appendix B.3. Among all regularization weights, $\lambda$ had the most noticeable effect, since it controls the relative importance of the MMD loss and therefore the strength of posterior decoupling at the distribution level. In our experiments, decreasing $\lambda$ was the only hyperparameter change that consistently affected performance.

The batch size was set to 128, and training was run for a maximum of 500 epochs. The temperature $\tau$ was initialized to 1 annealed during training, and early stopping was considered only after $\tau < 0.5$. After this point, training was terminated when both losses $\mathcal{L}_x$ and $\mathcal{L}_y$ stabilized and cluster assignments converged.

### A.5.3. CASL-VAE Algorithm

Detailed training procedure of CASL-VAEis disclosed by Algorithm 1.

**Algorithm 1** CASL-VAE training procedure.

1: **Input:** Target samples $\mathcal{X}$, reference samples $\mathcal{Y}$; encoders $e_{\phi_z}, e_{\phi_s}$; decoders $d_{\theta_z}, d_{\theta_s}$; batch size $B$; initial temperature $\tau_0$; minimum temperature $\tau_{\min}$; decay rate $r$; maximum steps $S$.
2: **Initialize:** $\tau \leftarrow \tau_0$, learnable prior mean embedding parameters $\psi$
3: **Learned prior:** $p(s_{\text{cont}} \mid s_{\text{cat}}) = \mathcal{N}(\mu_\psi(s_{\text{cat}}), I)$
4: **for** $step = 1$ **to** $S$ **do**
5:   **if** stopping criterion is met **then**
6:     **break**
7:   **end if**
8:   **if** $step \bmod 100 = 0$ **then**
9:     $\tau \leftarrow \max(\tau_{\min}, \tau_0 \exp(-r \cdot step))$
10:   **end if**
11:   Sample minibatches $\{x_b\}_{b=1}^B \sim \mathcal{X}$ and $\{y_b\}_{b=1}^B \sim \mathcal{Y}$
12:   **Encode common latent:**

$$(\mu_z(x_b), \sigma_z(x_b)) = e_{\phi_z}(x_b), \quad z_x \sim \mathcal{N}\big(\mu_z(x_b), \sigma_z^2(x_b) I\big)$$

$$(\mu_z(y_b), \sigma_z(y_b)) = e_{\phi_z}(y_b), \quad z_y \sim \mathcal{N}\big(\mu_z(y_b), \sigma_z^2(y_b) I\big)$$

13:   **Encode salient latents for target samples:**

$$\text{proj}_x = e_{\phi_s}^{\text{embed}}(x_b)$$

$$\alpha(x_b) = e_{\phi_s}^{\text{cat}}(\text{proj}_x) \quad \Rightarrow \quad s_{\text{cat},x} \sim \text{Gumbel-Softmax}(\alpha(x_b), \tau)$$

$$\mu_s(s_{\text{cat},x}, x_b) = e_{\phi_s}^{\text{cont}}\big(\text{concat}(s_{\text{cat},x}, \text{proj}_x)\big) \quad \Rightarrow \quad s_{\text{cont},x} \sim \mathcal{N}\big(\mu_s(s_{\text{cat},x}, x_b), I\big)$$

14:   Fix salient latents for reference samples to preset values:

$$s'_y = \text{concat}(s'_{\text{cat},y} = 1,\ s'_{\text{cont},y} = 0)$$

15:   Decode reconstructions:

$$\hat{x} = d_{\theta_z}(z_x) + d_{\theta_s}(\text{concat}(s_{\text{cat}_x}, s_{\text{cont}_x}))$$

$$\hat{y} = d_{\theta_z}(z_y) + d_{\theta_s}(s'_y)$$

16:   Compute ELBO terms $\mathcal{L}_x$, $\mathcal{L}_y$ (Eqs. 12, 13), using the learned prior $p(s_{\text{cont}} \mid s_{\text{cat}})$
17:   Compute regularizers $\mathcal{R}_{\text{MMD}}$, $\mathcal{R}_{\text{LC}}$, $\mathcal{R}_{\text{BCE}}$, $\mathcal{R}_{\text{ref}}$ (Section 3.6)
18:   Define objective $\mathcal{J} = -\mathcal{L}_x - \mathcal{L}_y + \lambda \mathcal{R}_{\text{MMD}} + \eta \mathcal{R}_{\text{LC}} + \zeta \mathcal{R}_{\text{BCE}} + \gamma \mathcal{R}_{\text{ref}}$
19:   Update parameters $\{\phi_z, \phi_s, \theta_z, \theta_s, \psi\}$ using ADAM by minimizing $\mathcal{J}$
20: **end for**
21: **Return:** trained encoders, decoders, prior mean embedding $\mu_\psi$

# B. Appendix: Experiments

## B.1. Brain Regions Included in Target Subtype Patterns for the Semi-synthetic Experiment

Table 5 lists the brain regions (bilateral left and right components, yielding two ROIs per region) included in the three subtype patterns (S1–S3) used to construct the semi-synthetic datasets.

*Table 5.* Brain regions included in the semi-synthetic disease patterns used in this study. Each subtype (S1–S3) affects 7 bilateral brain regions (left and right hemispheres; 14 ROIs in total), with partial overlap across subtypes. OP: opercular part.

| Brain Region (bilateral) | S1 | S2 | S3 |
|---|---|---|---|
| Amygdala | ✓ | | |
| Temporal pole | ✓ | ✓ | ✓ |
| Hippocampus | ✓ | ✓ | |
| Angular gyrus | | | ✓ |
| Entorhinal area | ✓ | | |
| Frontal operculum | | ✓ | |
| Inferior temporal gyrus | ✓ | | |
| Lateral orbital gyrus | | | ✓ |
| Medial frontal cortex | | ✓ | ✓ |
| Middle frontal gyrus | | ✓ | |
| Middle occipital gyrus | | | ✓ |
| OP of inferior frontal gyrus | | ✓ | |
| Parahippocampal gyrus | ✓ | | |
| Posterior insula | | ✓ | |
| Parietal operculum | ✓ | | ✓ |
| Supramarginal gyrus | | | ✓ |

## B.2. Experiment Procedure

For fair comparison in the semi-synthetic experiments, all ROI volume data were first residualized for age, sex, and intracranial volume using only the training reference data. The residuals were then standardized using the mean and standard deviation of the training reference data. The same preprocessing pipeline was applied to all baseline methods (K-means, GMM, GMVAE, cVAE, SepVAE, SmileGAN; refer Section 4.1).

CASL-VAE uses a common latent space of dimension $k = 15$, a single discrete salient latent with $M$ categories, and a continuous salient latent of dimension $t = 2$ (Appendix A.5.1). To ensure comparable model capacity across VAE-based methods, we fixed the total latent dimensionality to 18 for all models. Output layers were linear with Gaussian likelihood. The baselines were configured to match this total capacity while respecting their original design:

- **cVAE**: a single unified latent space of dimension 18.
- **SepVAE**: common latent dimension 15 and salient latent dimension 3. SepVAE was implemented with a similar encoder–decoder architecture as CASL-VAE, but without a hierarchical salient latent structure.
- **GMVAE**: a single discrete mixture component with $M$ categories and a continuous within-component latent dimension 17. To prevent mixture collapse and ensure subtype separation, we used identity covariance and learned component-specific means conditioned on the discrete mixture variable, following (Shu et al., 2017).

**Model configurations:** We evaluate clustering under two settings: (i) target-only clustering and (ii) target+reference clustering. K-means and GMM were applied to both settings. GMVAE is an unsupervised deep clustering model and was therefore applied only to target-only clustering. Smile-GAN is a semi-supervised deep clustering method that uses a Generative Adversarial Network to identify target subtypes by contrasting against the reference domain; thus it can be used for target-only clustering but not for target+reference clustering. cVAE and SepVAE are generative models trained on both target and reference data, with SepVAE additionally enforcing separation of reference and target samples in its salient latent space. To assess clustering performance for these models, we apply K-means to their inferred latent representations (unified latent for cVAE, salient latent for SepVAE).

**Training and evaluation:** All models were trained using the Adam optimizer (Kingma & Ba, 2015) with a learning rate of $10^{-3}$ and a batch size of 128. Models were trained 10 times using repeated hold-out cross-validation on unpaired reference and target samples. In each repetition, 80% of the data was used for training and the remaining 20% of the data was held out for assessing model performance. Reported results correspond to the mean and standard deviation or 95% confidence intervals (computed assuming a Student's t-distribution) across repetitions.

Clustering performance and reconstruction quality were evaluated using the Adjusted Rand Index (ARI) and mean absolute error (MAE), respectively, computed on the held-out validation set for each repetition. For conditional paired-sample generation, we additionally evaluated on a separate semi-synthetic test set containing ground-truth paired reference–target samples. Specifically, for each training repetition, the trained model generated paired reference domain samples for held-out pseudo-target samples, and cosine similarity was computed between the generated and true groud-truth paired reference samples.

GMVAE does not support paired sample generation because it was implemented for target-only clustering and does not model the reference domain, so it was excluded from this evaluation. For cVAE, paired samples were generated by encoding pseudo-target samples into the latent representation and decoding them conditioned on the reference domain label. For SepVAE, the common latent was inferred from the pseudo-target samples, while the salient latent was fixed to an uninformative vector **0** during decoding. CASL-VAE follows the same, except the uninformative vector is set as $(s'_{\text{cat}} = 1, s'_{\text{cont}} = 0)$ (refer Eq. 16).

### B.3. Ablation Analysis

To assess model sensitivity to effect size of target-specific variation, we trained CASL-VAE on semi-synthetic unpaired reference-target data where the pseudo-target samples were simulated with extremely mild (0-10%), mild (10-20%), and moderate (20-30%) atrophy rates. We also assessed robustness of model performance to hyperparameter-selection. For transaparency, we also performed this on the dataset simulated with the typical atrophy rate (10-30%).

For each semi-synthetic dataset, we trained CASL-VAE with varying hyper-parameters. Specifically, we changed each parameter from 0% (no regularization) to 200% of the preset hyperparameter value (defined in Appendix A.5.2), while keeping all other parameters fixed to the preset values. For each of these ablations, we trained the model 10 times using repeated hold-out cross-validation analysis described in Appendix B.2, and reported the average model performance on the held-out set per training loop.

Figure 5(a) reports the Adjusted Rand Index (ARI) for target-only clustering, while Figure 5(b) reports ARI for joint target+reference clustering. ARI was computed between the predicted cluster assignments and the ground-truth subtype labels in the semi-synthetic experiments.

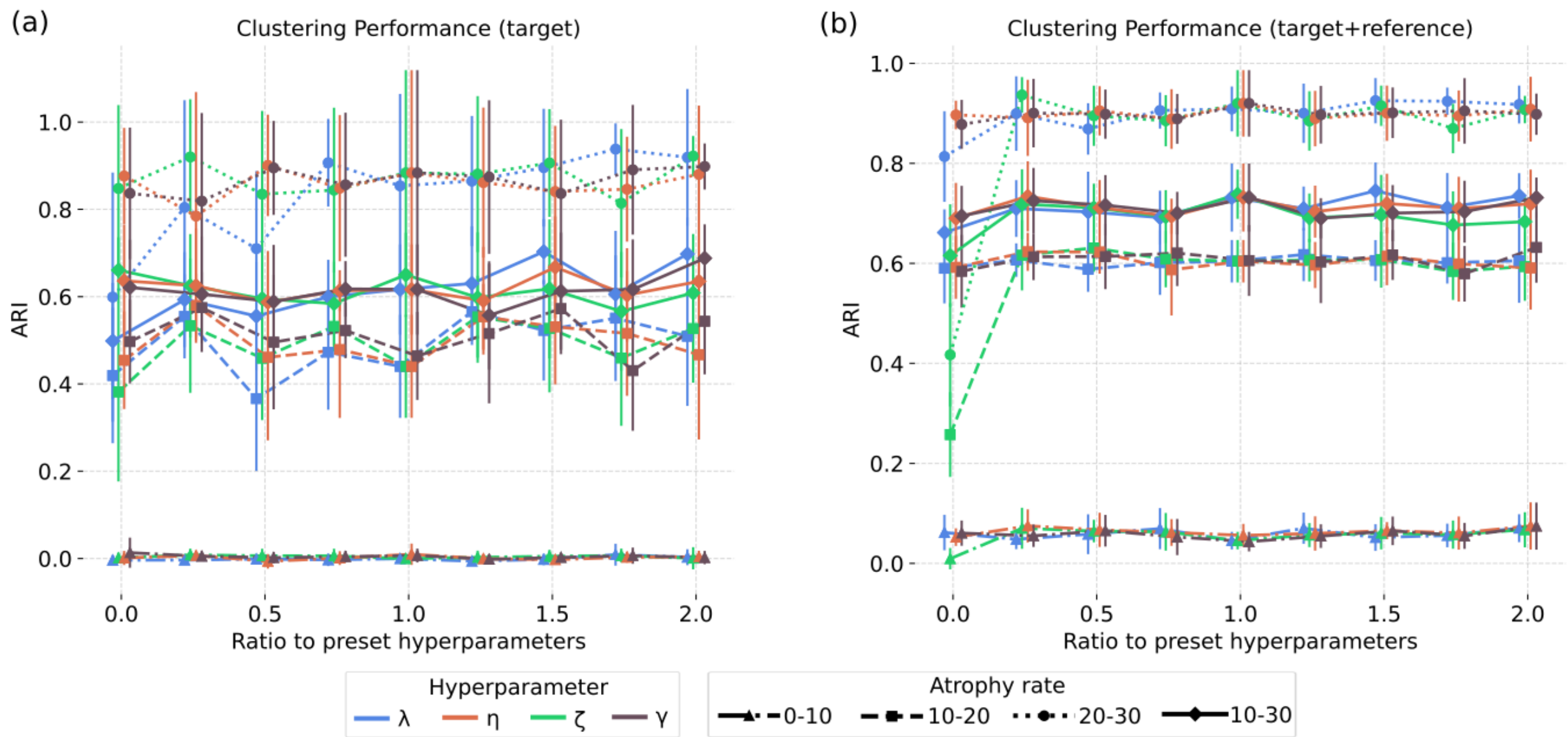

*Figure 5.* Ablation analysis of CASL-VAE clustering performance under varying regularization strengths and atrophy rates. (a) ARI for target-only clustering and (b) ARI for target+reference clustering, evaluated across 10 repeated hold-out validation splits. Curves show mean ± standard deviation across repetitions. Each regularization weight is scaled relative to its default value, with ratios ranging from 0 to 2 in increments of 0.25. Here, $\lambda$ controls $\mathcal{R}_{\text{MMD}}$, $\eta$ controls $\mathcal{R}_{\text{LC}}$, $\zeta$ controls $\mathcal{R}_{\text{BCE}}$, and $\gamma$ controls $\mathcal{R}_{\text{ref}}$.

We observe that CASL-VAE's ability to recover target subtypes degrades rapidly under extremely mild atrophy (0–10%), with ARI (target) approaching zero, indicating near-random cluster assignment. This suggests that when the effect size of target-specific variation is very small, reliable subtype discovery becomes challenging. This behavior is expected in a contrastive setting: when target-specific effects are very small, the target distribution substantially overlaps with the reference distribution, making reliable subtype discovery inherently challenging. As the effect size increases, target-specific variation becomes more distinguishable from the reference domain, and clustering performance improves consistently. A similar trend is observed for target+reference clustering, where stronger target-specific effects lead to clearer separation between reference samples and target subtypes.

Ablation results further highlight the role of individual regularization terms. Removing the MMD regularization on the common latent space leads to a noticeable drop in ARI, indicating degraded domain-invariant representation learning of the common latent. This suggests that, without MMD, target-specific variation may partially leak into the common latent space, weakening the separation between shared and salient factors. Likewise, removing the BCE loss substantially reduces performance in separating reference samples from target subtypes. Importantly, once the regularization terms are active, CASL-VAE exhibits stable performance across a broad range of hyperparameter strengths. Performance degrades primarily when specific regularization terms are removed entirely, demonstrating both the necessity of these components and robustness to precise hyperparameter tuning.

### B.4. Results from semi-synthetic data analysis

Additional results from the semi-synthetic experiment measuring reconstruction quality.

*Table 6.* Reconstruction performance (MAE) on pseudo-PT (target domain) samples in semi-synthetic data. Values show mean ± std with 95% CI. Lower is better.

| Model | MAE ↓ | |
|---|---|---|
| | 10–20% | 10–30% |
| cVAE | 0.6355 ± 0.0079 (0.6295, 0.6415) | 0.6319 ± 0.0089 (0.6252, 0.6386) |
| SepVAE | 0.6118 ± 0.0076 (0.6061, 0.6175) | 0.6101 ± 0.0081 (0.6040, 0.6162) |
| CASL-VAE (Ours) | **0.6094 ± 0.0116 (0.6006, 0.6181)** | **0.6070 ± 0.0092 (0.6001, 0.6139)** |

*Table 7.* Reconstruction performance (MAE) on CN (reference domain) samples in semi-synthetic data. Values show mean ± std with 95% CI. Lower is better.

| Model | MAE ↓ | |
|---|---|---|
| | 10–20% | 10–30% |
| cVAE | 0.6394 ± 0.0087 (0.6328, 0.6460) | 0.6399 ± 0.0078 (0.6340, 0.6458) |
| SepVAE | **0.6146 ± 0.0098 (0.6073, 0.6220)** | **0.6151 ± 0.0092 (0.6082, 0.6221)** |
| CASL-VAE (Ours) | 0.6240 ± 0.0121 (0.6149, 0.6331) * | 0.6219 ± 0.0095 (0.6147, 0.6291) * |

**Note:** *MAE of CASL-VAE is not significantly different from SepVAE (95% CIs overlap).

### B.5. Real-world data analysis and association of target-specific variation with clinical variables

For each test sample $i$, CASL-VAE yields an interpretable latent representation $(z_i, s_{\text{cat},i}, s_{\text{cont},i})$. Target-specific variation is quantified by decoding the salient latent and subtracting the reference baseline:

$$\Delta_i = d_{\theta_s}(s_{\text{cat},i}, s_{\text{cont},i}) - d_{\theta_s}(s'_{\text{cat}}, s'_{\text{cont}}). \tag{33}$$

This isolates deviations relative to the reference domain and provides a direct representation of target-specific structure in the input space. To obtain a scalar measure of overall deviation, we compute a deviation index (DI) as the mean absolute value of $\Delta_i$ across all dimensions:

$$\text{DI}_i = \frac{1}{D}\sum_{d=1}^{D} |\Delta_{i,d}| \, ,$$

where $D$ is the dimensionality of the decoded output. Another way of interpreting this is that DI measures the overall magnitude of an individual's deviation from their paired reference.

To evaluate whether CASL-VAE captures meaningful disease-related variation in real data, we correlated the deviation index (DI) with clinical variables and biomarkers reflecting AD pathology severity. For each clinical variable, we only selected participants who had the variable available to include in the statistical analysis. The clinical variables used were ADNI-MEM, ADNI-LAN, and ADNI-EF, which are composite scores measuring memory, language, and executive function, respectively (higher values indicate better performance). MMSE is a global cognitive score (higher values indicate better cognition), and ADAS-Cog-13 is another complementary score that measures cognitive performance across 13 tasks, where higher scores indicate worse cognitive performance. APOE E4 carrier status indicates whether the participant carries at least one APOE E4 allele (0 = non-carrier, 1 = carrier), with APOE E4 being a genetic risk factor for AD. We also included cerebrospinal fluid (CSF) biomarkers: CSF-Abeta (amyloid-$\beta$ 1–42; lower values indicate greater amyloid pathology), CSF-pTau (phosphorylated tau; higher values indicate greater tau pathology), and CSF-Tau (total tau; higher values indicate greater neurodegeneration). For dichotomous variables (APOE E4 carrier), point-biserial correlation was used; Pearson correlation was used for continuous variables.

*Table 8.* Correlation of clinical and biomarker variables with the overall deviation index (DI). DI measures the mean absolute target-specific salient variation in each sample relative to the reference domain. Variables with $p < 0.001$ are marked accordingly.

| **Variable** | **Correlation with DI** | **p-value** | $p < 0.001$ |
|---|---|---|---|
| ADNI-MEM | $-0.469$ | $9.99 \times 10^{-37}$ | Yes |
| ADNI-LAN | $-0.348$ | $6.85 \times 10^{-20}$ | Yes |
| ADNI-EF | $-0.276$ | $9.26 \times 10^{-13}$ | Yes |
| ADAS-COG-13 | 0.473 | $4.76 \times 10^{-37}$ | Yes |
| APOE E4 Carrier | 0.242 | $3.81 \times 10^{-10}$ | Yes |
| MMSE | $-0.354$ | $1.20 \times 10^{-20}$ | Yes |
| CSF-Abeta | $-0.254$ | $3.18 \times 10^{-6}$ | Yes |
| CSF-PTau | 0.146 | $8.24 \times 10^{-3}$ | No |
| CSF-Tau | 0.115 | $3.67 \times 10^{-2}$ | No |

Table 8 reports the results of this analysis. Higher DI values were associated with lower memory, language, and executive function scores (ADNI-MEM: $r = -0.469$, ADNI-LAN: $r = -0.348$, ADNI-EF: $r = -0.276$), higher overall cognitive impairment (ADAS-Cog-13: $r = 0.473$), lower MMSE scores ($r = -0.354$), and increased likelihood of being an APOE E4 carrier ($r = 0.242$). CSF Abeta showed a negative correlation ($r = -0.254$), while correlations with CSF Tau and pTau were weak and not significant. These results indicate that the salient latent variables in CASL-VAE capture clinically meaningful target-specific variation across samples.

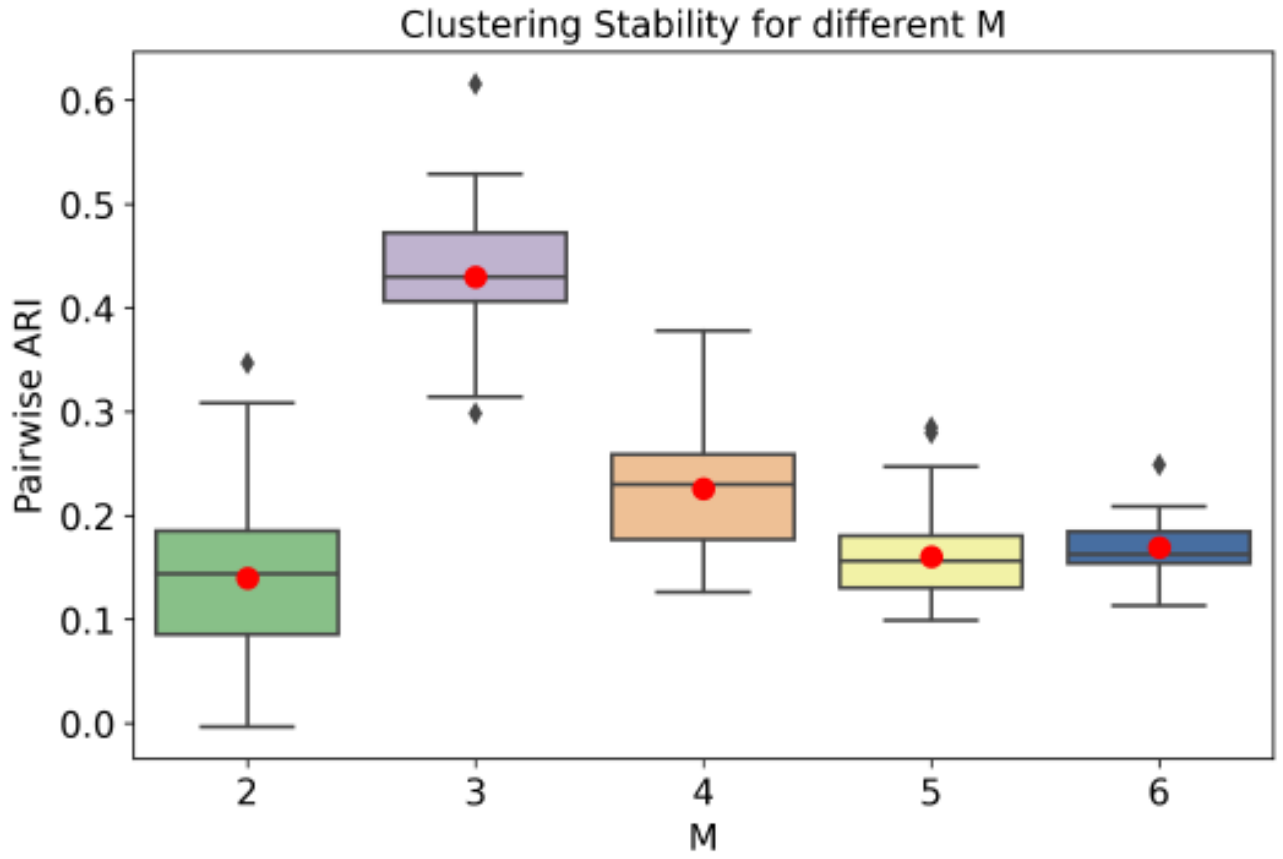

*Figure 6.* Clustering stability of CASL-VAE on ADNI2/GO training data. Box-plots show pairwise ARI between clusterings from 10 folds of repeated hold-out cross-validation, for $M = 2, \ldots, 6$.